\documentclass{article}

\usepackage{amsmath,amsfonts,amssymb,times,graphicx,natbib,algorithm,algorithmic,hyperref}

\usepackage{booktabs} 
\usepackage[T1]{fontenc}    
\usepackage{booktabs}       
\usepackage{nicefrac}       
\usepackage{microtype}      

\usepackage[accepted]{hill2019}

\icmltitlerunning{
Teaching AI to Explain its Decisions Using Embeddings and Multi-Task Learning
}

\newcommand{\ignore}[1]{}

\newcommand{\ted}{TED }

\begin{document}

\twocolumn[
\icmltitle{
Teaching AI to Explain its Decisions\\
Using Embeddings and Multi-Task Learning
}


\icmlsetsymbol{equal}{*}

\begin{icmlauthorlist}
\icmlauthor{Noel C.\ F.\ Codella}{equal,ibm}
\icmlauthor{Michael Hind}{equal,ibm}
\icmlauthor{Karthikeyan Natesan Ramamurthy}{equal,ibm}
\icmlauthor{Murray Campbell}{ibm}
\icmlauthor{Amit Dhurandhar}{ibm}
\icmlauthor{Kush R.\ Varshney}{ibm}
\icmlauthor{Dennis Wei}{ibm}
\icmlauthor{Aleksandra Mojsilovi\'c}{ibm}
\end{icmlauthorlist}

\icmlaffiliation{ibm}{IBM Research AI, Yorktown
Heights, NY, USA.}

\icmlcorrespondingauthor{Michael Hind}{hindm@us.ibm.com}

\icmlkeywords{explainability, interpretability, transparency}

\vskip 0.3in
]


\printAffiliationsAndNotice{\icmlEqualContribution}

\begin{abstract}
Using machine learning in high-stakes applications 
often requires predictions to be
accompanied by explanations comprehensible to the domain user, who
has ultimate responsibility for decisions and outcomes.  
Recently, a new framework for providing explanations, called
TED~\cite{ted-vision2019}, 
has been proposed to provide meaningful explanations for predictions.
This framework augments training data to include explanations elicited
from domain users, in addition to features and labels.
This approach ensures that explanations for predictions are tailored to the
complexity expectations and domain knowledge of the consumer.  

In this paper, we build on this foundational work, by exploring more
sophisticated instantiations of the TED 
framework and empirically evaluate their effectiveness in two diverse
domains, chemical odor and skin cancer prediction.
Results demonstrate that meaningful explanations can be reliably
taught to machine learning algorithms, and in some cases, improving
modeling accuracy. 

\ignore{
A joint model is then learned to produce both labels and explanations from the input features. 
Evaluation spans multiple modeling techniques on an image dataset and
a chemical odor dataset, showing that our approach is generalizable
across domains and algorithms. 
}
\end{abstract}

\section{Introduction}
\label{sec:intro}


New regulations call for automated decision making systems to provide
``meaningful information'' on the logic used to reach conclusions
\citep{gdpr-goodman,gdpr-wachter,SelbstP2017}. \citet{SelbstP2017}
interpret the concept of ``meaningful information'' as information
that should be understandable to the audience (potentially individuals
who lack specific expertise), is actionable, and  is flexible enough
to support various technical approaches.

Recently, \citet{ted-vision2019} introduced a new framework, called \ted ({\em Teaching Explanations for Decisions}),  for
providing meaningful explanations for machine learning predictions.
The framework requires the training dataset to include an
explanation ($E$), along with the features ($X$) and target label ($Y$).
A model is learned from this training set so that
predictions include a label and an explanation, both of which are from
the set of possible labels ($Y$) and explanations ($E$) that are
provided in the training data.

This approach has several advantages. The explanations provided should
meet the complexity capability and domain knowledge of the target user
because they are providing the training explanations.  The explanations should be as
accurate as the underlying predictions.  The framework is general in
that it can be applied to any supervised machine learning algorithm.

In addition to describing the framework and other advantages,
\citet{ted-vision2019} also describe a simple instantiation of the
framework, based on the Cartesian product, that combines the
$Y$ label and $E$ explanation to form a new training dataset ($X$,
$YE$), which is then fed into any machine learning algorithm.  The
resulting model will produce predictions ($YE$) that can be
decomposed to the $Y$ and $E$ component.  This instantiation is
evaluated on two synthetic datasets for playing tic-tac-toe and
loan repayment.  Their results show
``{\em (1) To the extent that user explanations follow simple logic, very
high explanation accuracy can be achieved; (2) 
Accuracy in predicting $Y$ not only does not suffer but actually improves.}''

In this work, we explore more sophisticated instantiations of the \ted
framework that leverage training explanations to improve the accuracy
of prediction as well as the generation of  explanations.
Specifically, we explore  bringing together the labels and explanations in a multi-task
setting, as well as building upon the
tradition of similarity metrics, case-based reasoning and
content-based retrieval.   

Existing approaches that only have access to features and labels are
unable to find \emph{meaningful} similarities. However, with the
advantage of having training features, labels, \emph{and}
explanations, we propose to learn feature embeddings  
guided by labels and explanations.  This allows us to infer
explanations for new data using nearest neighbor approaches. We
present a new objective function to learn an embedding to optimize
$k$-nearest neighbor ($k$NN) search for both prediction accuracy as
well as holistic human relevancy to enforce that returned neighbors
present meaningful information. The proposed embedding approach is
easily portable to a diverse set of label and explanation spaces
because it only requires a notion of similarity between examples in
these spaces. Since any predicted explanation or label is obtained
from a simple combination of training examples, \textit{complexity}
and \textit{domain} match is achieved with no further effort. We also
demonstrate the multi-task instantiation wherein labels and
explanations are predicted together from features. In contrast to the
embedding approach, we need to change the structure of the ML model 
for this method due to the modality and type of the label and
explanation space. 

We demonstrate the proposed paradigm using the two instantiations on
publicly-available 
olfactory pleasantness dataset
\citep{olfs} and melanoma classification dataset
\citep{codella2018skin}.\footnote{An extended version of this paper~\cite{ted-knn-arxiv2018}
also evaluates an image aesthetics dataset \citep{kong2016aesthetics}.}
Teaching explanations requires a
training set that contains explanations.  Since such datasets are not
readily available, we use the attributes given with the
pleasantness dataset in a unique way: as collections of meaningful
explanations. For the melanoma classification dataset, we will use the
groupings given by human users described in
\citet{codella2018collaborative} as the explanations. 

The main contributions of this work are:
\begin{itemize}

\ignore{
\item A new approach for machine learning algorithms to provide \emph{meaningful} explanations that match the complexity and domain of consumers by eliciting training explanations directly from them.  We name this paradigm TED for `Teaching Explanations for Decisions.'
}

\item Instantations and  evaluation of several candidate \ted approaches, some that learn efficient embeddings that can be used to infer labels and explanations for novel data, and some that use multi-task learning to predict labels and explanations.




\item Evaluation on disparate datasets with diverse label and explanation spaces demonstrating the efficacy of the paradigm.
\end{itemize}

\section{Related Work} 
\label{sec-related}

Prior work in providing explanations can be partitioned into several areas:

\begin{enumerate}
\item \label{interp}
Making existing or enhanced models {\em interpretable}, i.e.\ to provide a precise description of how the model determined its
decision (e.g.,~\citet{RibieroSG2016, montavon2017methods, unifiedPI}).

\item \label{2nd-model}
Creating a simpler-to-understand model, such as a small number of logical expressions, that mostly 
matches the decisions of the deployed model (e.g.,~\citet{bastani2017interpreting, Caruana:2015}), or directly training a simple-to-understand model from the data (e.g.,~\citet{dash2018boolean, JMLR:v18:16-003, cohen1995fast, breiman2017classification}).

\item \label{rationale}
Leveraging ``rationales'', ``explanations'', ``attributes'', or other ``privileged information'' in the training data 
to help improve the accuracy of the
algorithms (e.g.,~\citet{sun-dejong-2005,zaidan-eisner:2008:gen,rationales,peng-vision}). 

\item \label{gen-rationale}
Work in the natural language processing and computer vision domains that generate rationales/explanations derived from input text (e.g., \citet{lei2016rationalizing,Yessenalina:2010:AGA:1858842.1858904,hendricks-2016}).

\item \label{cbir}
Content-based retrieval methods that provide explanations as {\em evidence} employed for a prediction, i.e. $k$-nearest neighbor classification and regression (e.g., \citet{cbir3,cbir4,cbir5,patientsimilarity}).

\end{enumerate}

The first two groups attempt to precisely describe how a
machine learning decision was made, which is particularly relevant for
AI system builders.  This insight can be used to improve the AI 
system and may serve as the seeds for an explanation to a non-AI expert.
However, work still remains to determine if these seeds are sufficient
to satisfy the needs of a non-AI expert.  In particular, when the underlying features are not human comprehensible, these approaches are inadequate for providing human consumable explanations.

The third group, like this work, leverages additional information (explanations) in the training data, but with different goals.  The third group uses the explanations to create a more accurate model; the \ted framework leverages the explanations to teach how to generate explanations for new predictions.  

The fourth group seeks to generate textual explanations with predictions. For text classification, this involves selecting the minimal necessary content from a text body that is sufficient to trigger the classification. For computer vision~\citep{hendricks-2016}, this involves utilizing textual captions to automatically generate new textual captions of images that are both descriptive as well as discriminative. While serving to enrich an understanding of the predictions, these systems do not necessarily facilitate an improved ability for a human user to understand system failures.  

The fifth group creates explanations in the form of {\em decision evidence}: using some feature embedding to perform {\em k}-nearest neighbor search, using those {\em k} neighbors to make a prediction, and demonstrating to the user the nearest neighbors and any relevant information regarding them. Although this approach is fairly straightforward and holds a great deal of promise, it has historically suffered from the issue of the semantic gap: distance metrics in the realm of the feature embeddings do not necessarily yield neighbors that are relevant for prediction. More recently, deep feature embeddings, optimized for generating predictions, have made significant advances in reducing the semantic gap. However, there still remains a ``meaning gap'' --- although systems have gotten good at returning neighbors with the same label as a query, they do not necessarily return neighbors that agree with any {\em holistic human measures} of similarity. As a result, users are not necessarily inclined to trust system predictions.


\citet{doshiexpl} discuss the societal, moral, and legal expectations of AI explanations, provide
guidelines for the content of an explanation, and recommend that
explanations of AI systems be held to a similar standard as humans.  Our
approach is compatible with their view.   \citet{biran-cotton-2017} provide an excellent overview and taxonomy of explanations and justifications in machine learning.

\citet{miller2017explanation-review} and \citet{miller2017inmates}
argue that explainable AI solutions need to meet the needs of the
users, an area that has been well studied in philosophy, psychology,
and cognitive science.  They provides a brief survey of the most
relevant work in these fields to the area of explainable AI.
They, along with \citet{rsi}, call for more rigor
in this area.

\section{Methods}  \label{methods}

The primary motivation of the TED paradigm is to provide meaningful explanations to consumers by leveraging the consumers' knowledge of what will be meaningful to them.
Section~\ref{sec-problem} formally describes the problem space that
defines the TED approach. 
\ignore{
 One simple learning approach to this problem is to expand the label space to be the Cartesian product of the original labels and the provided explanations.  Although quite simple, this approach has a number of pragmatic advantages in that it is easy to incorporate, it can be used for any learning algorithm, it does not require any changes to the learning algorithm, and does not require owners to make available their algorithm.  It also has the possibility of some indirect benefits because requiring explanations will improve auditability (all decisions will have explanations) and potentially reduce bias in the training set because inconsistencies in explanations may be discovered.
}

This paper focuses on  instantiations of the TED approach that
leverages the explanations to improve model prediction and possibly
explanation accuracy.   
Section~\ref{jointpairwise} takes this approach  to learn feature embeddings and explanation embeddings in a joint and aligned way to permit neighbor-based explanation prediction.  It presents a new objective function to learn an embedding to optimize $k$NN search for both prediction accuracy as well as holistic human relevancy to enforce that returned neighbors present meaningful information. We also discuss multi-task learning in the label and explanation space as another instantiation of the TED approach, that we will use for comparisons.
 
\subsection{Problem Description} 
\label{sec-problem}
Let $X\times Y$ denote the input-output space, with $p(x,y)$ denoting the joint distribution over this space, where $(x,y)\in X\times Y$. Then typically, in supervised learning one wants to estimate $p(y|x)$.

In our setting, we have a triple $X\times Y\times E$ that denotes the input space, output space, and explanation space, respectively. We then assume that we have a joint distribution $p(x,y,e)$ over this space, where $(x,y,e)\in X\times Y\times E$. In this setting we want to estimate $p(y,e|x)=p(y|x)p(e|y,x)$. Thus, we not only want to predict the labels $y$, but also the corresponding explanations $e$ for the specific $x$ and $y$ based on historical explanations given by human experts.
The space $E$ in most of these applications is quite different than $X$ and 
has similarities with $Y$ in that it requires human judgment.

We provide methods to solve the above problem. Although these methods can be used even when $X$ is human-understandable, we envision the most impact for applications where this is not the case, such as the olfaction dataset described in Section~\ref{sec-eval}.

\subsection{Candidate Approaches}
\label{jointpairwise}

We propose several candidate implementation approaches to teach labels and explanations from  training data, and predict them for unseen test data. We describe the baseline regression and embedding approaches. The particular parameters and specific instantiations are provided in Section \ref{sec-eval}.

\subsubsection{Baseline for Predicting $Y$ or $E$}
To set the baseline, we trained a regression (classification) network
on the datasets to predict $Y$ from $X$ using the mean-squared error
(cross-entropy) loss. This cannot be used to infer $E$ for a novel
$X$. A similar learning approach was used to predict $E$ from $X$. If $E$ is vector-valued, we used multi-task learning.

\subsubsection{Multi-task Learning to Predict $Y$ and $E$ Together}
We trained a multi-task network to predict $Y$ and $E$ together from $X$. Similar to the previous case, we used appropriate loss functions.

\subsubsection{Embeddings to Predict $Y$ and $E$}
We propose to use the activations from the last fully connected hidden layer of the network trained to predict $Y$ or $E$ as embeddings for $X$. Given a novel $X$, we obtain its $k$ nearest neighbors in the embedding space from the training set, and use the corresponding $Y$ and $E$ values to obtain predictions as weighted averages. The weights are determined using a Gaussian kernel on the distances in the embedding space of the novel $X$ to its neighbors in the training set. This procedure is used with all the $k$NN-based prediction approaches.

\subsubsection{Pairwise Loss for Improved Embeddings}
Since our key instantiation is to predict $Y$ and $E$ using the $k$NN approach described above, we propose to improve upon the embeddings of $X$ from the regression network by explicitly ensuring that points with similar $Y$ and $E$ values are mapped close to each other in the embedding space. 
For a pair of data points $(a,b)$ with inputs $(x_a,x_b)$, labels
$(y_a,y_b)$, and explanations $(e_a,e_b)$, we define the following
pairwise loss functions for creating the embedding $f(\cdot)$, where
the shorthand for $f(x_i)$ is $f_i$ for clarity below: 
\vspace{-2ex}
\begin{multline}\label{equation-loss-x-y}
L_{x,y}(a,b)\\ = \begin{cases}
1 - \cos(f_a,f_b), & ||y_{a} - y_{b}||_1 \leq c_1,\\ 
\max(\cos(f_a,f_b) - m_1, 0), & ||y_{a} - y_{b}||_1 > c_2,
\end{cases}
\end{multline}
\vspace{-2ex}
\begin{multline}\label{equation-loss-x-e}
L_{x,e}(a,b)\\ = \begin{cases}
1 - \cos(f_a,f_b), & ||e_{a} - e_{b}||_1 \leq c_3,\\ 
\max(\cos(f_a,f_b) - m_2, 0), & ||e_{a} - e_{b}||_1 > c_4.
\end{cases}
\end{multline}
The cosine similarity $\cos(f_a,f_b)=\frac{f_a\cdot f_b}{||f_a||_2||f_b||_2}$, where $\cdot$ denotes the dot product between the two vector embeddings and $||.||_p$ denotes the $\ell_p$ norm. Eqn.~(\ref{equation-loss-x-y}) defines the embedding loss based on similarity in the $Y$ space. If $y_a$ and $y_b$ are close, the cosine distance between $x_a$ and $x_b$ will be minimized. If $y_a$ and $y_b$ are far, the cosine similarity will be minimized (up to some margin $m_1 \geq 0$), thus maximizing the cosine distance. It is possible to set $c_2 > c_1$ to create a clear buffer between neighbors and non-neighbors. 
The loss function (\ref{equation-loss-x-e}) based on similarity in the $E$ space is exactly analogous. We combine the losses using $Y$ and $E$ similarities as
\begin{equation} \label{equation-loss-x-y-e}
L_{x,y,e}(a,b) = L_{x,y}(a,b) + w \cdot L_{x,e}(a,b),
\end{equation} where $w$ denotes the scalar weight on the $E$ loss. We set $w \leq 1$ in our experiments. The neighborhood criteria on $y$ and $e$ in (\ref{equation-loss-x-y}) and (\ref{equation-loss-x-e}) are only valid if they are continuous valued. If they are categorical, we will adopt a different neighborhood criteria, whose specifics are discussed in the relevant experiment below.

\section{Evaluation} \label{sec-eval}
To evaluate the \ted instantiations presented in this work, we focus on two fundamental questions:
\begin{enumerate}
\item Does the instantiation provide useful explanations? \label{ques-explanation}
\item How is the prediction accuracy impacted by incorporating explanations into the training? \label{ques-accuracy}
\end{enumerate}

Since the \ted approach can be incorporated into many kinds of learning algorithms, tested against many datasets, and used in many different situations, 
a definitive answer to these questions is beyond the scope of this paper.  Instead we try to address these two questions on two datasets, evaluating accuracy in the standard way.  

Determining if any approach provides useful explanations is a challenge and no  
consensus metric has yet to emerge~\citep{doshiexpl}.  However, the \ted approach has a unique advantage in dealing with this challenge.  Specifically, since it requires explanations be provided for the target dataset (training and testing),   one can evaluate the accuracy of a model's explanation ($E$) in a similar way that one evaluates the accuracy of a predicted label ($Y$). We provide more details on the metrics used in Section~\ref{sec-eval-metrics}.  In general, we expect several metrics of explanation efficacy to emerge, including those involving the target explanation consumers \citep{tip}.

\subsection{Datasets}
The \ted approach requires a training set that contains explanations.
Since such datasets are not readily available, we evaluate the
approach on 2 publicly available datasets in a unique way: 
Olfactory~\citep{olfs} and Melanoma detection~\citep{codella2018skin}.

\ignore{
The AADB (Aesthetics and Attributes
Database)~\citep{kong2016aesthetics} contains about $10,000$ images
that have been human rated for aesthetic quality ($Y \in [0, 1]$),
where higher values imply more aesthetically pleasing.  It also comes
with 11 attributes ($E$) that are closely related to image aesthetic
judgments by professional photographers. The attribute values are
averaged over 5 humans and lie in $[-1,1]$.  The training, test, and
validation partitions are provided by the authors and consist of
8,458, 1,000, and 500 images, respectively.
}

The Olfactory dataset~\citep{olfs}
is a challenge dataset describing various scents (chemical bondings and labels).
Each of the 476 rows represents a molecule with approximately $5000$ chemoinformatic features ($X$) 
(angles between bonds, types of atoms, etc.).  Each row also contains 21 human perceptions of the molecule, such as \emph{intensity},  \emph{pleasantness}, \emph{sour}, \emph{musky}, \emph{burnt}.  These are average values among 49 diverse individuals and lie in $[0, 100]$.
We take $Y$ to be the \emph{pleasantness} perception 
and $E$ to be the remaining 19 perceptions except for \emph{intensity}, since these 19 are known to be more fundamental semantic descriptors while pleasantness and intensity are holistic perceptions~\citep{olfs}. 
We use the standard training, test, and validation sets provided by the challenge organizers with $338$, $69$, and $69$ instances respectively.

\begin{figure}[t]
\centering
\includegraphics[width=8.5cm]{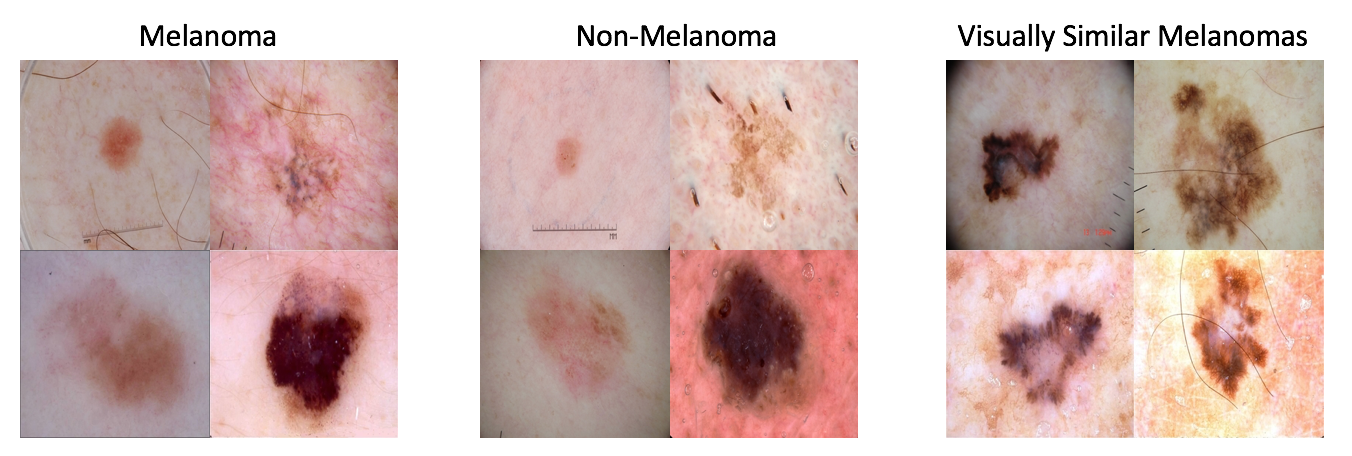}
\caption{Example images from the ISIC Melanoma detection dataset. The visual similarity between Melanoma and non-Melanoma images is seen from the left and middle images. In the right image, the visually similar lesions are placed in the same group (i.e., have the same $e$ value).}
\label{fig:melanoma_images}
\end{figure}

The 2017 International Skin Imaging Collaboration (ISIC) challenge on Skin Lesion Analysis Toward Melanoma Detection dataset \citep{codella2018skin} is a public dataset with $2000$ training and $600$ test images. Each image belongs to one of the three classes: melanoma (513 images), seborrheic keratosis (339 images) and benign nevus (1748 images). We use a version of this dataset described by \citet{codella2018collaborative}, where the melanoma images were partitioned to 20 groups, the seborrheic keratosis images were divided into 12 groups, and 15 groups were created for benign nevus, by a non-expert human user. We show some example images from this dataset in Figure \ref{fig:melanoma_images}. We take the $3$ class labels to be $Y$ and the $47$ total groups to be $E$. In this dataset, each $e$ maps to a unique $y$. We partition the original training set into a training set with $1600$ images, and a validation set with $400$ images, for use in our experiments. We continue using the original test set with $600$ images.

\subsection{Metrics}
\label{sec-eval-metrics}
An open question that we do not attempt to resolve here 
is the precise form that explanations should take. It is important that they 
match the mental model of the explanation consumer.  For example, one
may expect explanations to be categorical (as in loan approval reason
codes or our melanoma dataset) or discrete ordinal, as in human
ratings. Explanations may also be continuous in crowd sourced
environments, where the final rating is an (weighted) average over the
human ratings. This is seen in the
Olfactory
datasets that we consider, where each explanation is averaged over
49 individuals.

In the 
Olfactory dataset, since we use the existing continuous-valued attributes as explanations, we choose to treat them both as-is and discretized into $3$ bins, $\{-1, 0, 1\}$, representing negative, neutral, and positive values. The latter mimics human ratings (e.g.,~not pleasing, neutral, or pleasing).  
Specifically, we train on the original continuous $Y$ values and report absolute error (MAE) between $Y$ and a continuous-valued prediction $\hat{Y}$.  We also similarly discretize $Y$ and $\hat{Y}$ as $-1, 0, 1$.  We then report both absolute error in the discretized values (so that $\lvert 1 - 0 \rvert = 1$ and $\lvert 1 - (-1) \rvert = 2$) as well as $0$-$1$ error ($\hat{Y} = Y$ or $\hat{Y} \neq Y$), where the latter corresponds to conventional classification accuracy. 
We use bin thresholds of
$33.66$ and $49.68$ for Olfactory to partition the $Y$ scores in the training data into thirds. 

The explanations $E$ are treated similarly to $Y$ by computing $\ell_1$ distances (sum of absolute differences over attributes) before and after discretizing to $\{-1, 0, 1\}$.  
We do not, however, compute the $0$-$1$ error for $E$.
We use thresholds of
$2.72$ and $6.57$ for Olfactory, which roughly partitions the values into thirds based on the training data.

For the melanoma classification dataset, since both $Y$ and $E$ are categorical, we use classification accuracy as the performance metric for both $Y$ and $E$.

\begin{table*}[ht]
\begin{tiny}
\caption{Accuracy of predicting $Y$ and $E$ for  ISIC (left) and Olfactory (right) using different
methods (Section \ref{jointpairwise}). For ISIC, Baseline for $Y$ and
$E$ are classification networks. For Olfactory, Baseline LASSO and RF predict $Y$ from $X$. Multi-task LASSO
regression with $\ell_{21}$ regularization on the coefficient matrix
predicts Y\&E together, or just $E$. For both, Multi-task learning predicts both $Y$ and $E$ together, Embedding $Y$ + $k$NN uses the embedding from the last hidden layer of the baseline network that predicts $Y$. Pairwise $Y$ + $k$NN and Pairwise $E$ + $k$NN use the cosine embedding loss in 
(\ref{equation-loss-x-y}) and (\ref{equation-loss-x-e}) respectively to optimize the embeddings of $X$. Pairwise $Y$ \& $E$ + $k$NN uses the sum of cosine embedding losses in (\ref{equation-loss-x-y-e}) to optimize the embeddings of $X$.
}
\label{table-Olf-Mel}

\begin{tabular}{|c|c|c|c|c|} \hline
Algorithm	& $\lambda$ or K & Y Accuracy	& E Accuracy \\ \hline \hline
Baseline ($Y$) & NA & 0.7045 & NA   \\ \hline
Baseline ($E$) & NA & 0.6628 & 0.4107   \\ \hline \hline

 & 0.01 & 0.6711 & 0.2838  \\ \cline{2-4}
 & 0.1 & 0.6644 & 0.2838  \\ \cline{2-4}
 & 1 & 0.6544 & {\bf 0.4474}  \\ \cline{2-4}
Multi-task & 10 & 0.6778 & 0.4274  \\ \cline{2-4}
classification & 25 & {\bf 0.7145} & 0.4324  \\ \cline{2-4}
($Y$ \& $E$) & 50 & 0.6694 & 0.4057  \\ \cline{2-4}
 & 100 & 0.6761 & 0.4140  \\ \cline{2-4}
 & 250 & 0.6711 & 0.3957  \\ \cline{2-4}
 & 500 & 0.6327 & 0.3907  \\ \hline
 
 & 1 & 0.6962 & 0.2604  \\ \cline{2-4}
Embedding $Y$ & 2 & {\bf 0.6995} & 0.2604  \\ \cline{2-4}
+ & 5 & 0.6978 & 0.2604  \\ \cline{2-4}
$k$NN  & 10 & 0.6962 & 0.2604  \\ \cline{2-4}
 & 15 & 0.6978 & 0.2604  \\ \cline{2-4}
 & 20 & {\bf 0.6995} & 0.2604  \\ \hline

 & 1 & {\bf 0.6978} & 0.4357  \\ \cline{2-4}
Embedding $E$ & 2 & 0.6861 & 0.4357  \\ \cline{2-4}
+ & 5 & 0.6861 & 0.4357  \\ \cline{2-4}
$k$NN & 10 & 0.6745 & 0.4407  \\ \cline{2-4}
 & 15 & 0.6828 & {\bf 0.4374}  \\ \cline{2-4}
 & 20 & 0.6661 & 0.4424  \\ \hline 


& 1 & 0.7162 & 0.1619  \\ \cline{2-4}
Pairwise $Y$ & 2 & {\bf 0.7179} & 0.1619  \\ \cline{2-4}
+& 5 & {\bf 0.7179} & 0.1619  \\ \cline{2-4}
$k$NN & 10 & 0.7162 & 0.1619  \\ \cline{2-4}
& 15 & 0.7162 & 0.1619  \\ \cline{2-4}
& 20 & 0.7162 & 0.1619  \\ \hline


& 1 & 0.7245 & {\bf 0.3406}  \\ \cline{2-4}
Pairwise $E$ & 2 & 0.7279 & {\bf 0.3406}  \\ \cline{2-4}
+ & 5 & 0.7229 & 0.3389  \\ \cline{2-4}
$k$NN & 10 & 0.7279 & 0.3389  \\ \cline{2-4}
& 15 & {\bf 0.7329} & 0.3372  \\ \cline{2-4}
& 20 & 0.7312 & 0.3356  \\ \hline
\end{tabular}
\hfill
\begin{tabular}{|c|c|c|c|c|c|c|} \hline
	& 	& \multicolumn{3}{c|}{Performance on Y} & \multicolumn{2}{c|}{Performance on E}  \\ \cline{3-7}
	& 	&  & \multicolumn{2}{c|}{MAE} & \multicolumn{2}{c|}{MAE}    \\ \cline{4-7}
Algorithm	& $k$	& Class. Accuracy & Discretized & Continuous & Discretized &
Continuous \\ \hline \hline

Baseline LASSO ($Y$)     & NA &  0.4928 & 0.5072 & {\bf 8.6483} & NA & NA  \\ \hline
Baseline RF ($Y$)       & NA &  {\bf 0.5217} & {\bf 0.4783} & 8.9447 & NA & NA  \\
\hline \hline

Multi-task & & & & & & \\
 regression & NA & 0.4493 & 0.5507 & 11.4651 & 0.5034 &3.6536 \\ 
 ($Y$ \& $E$) & & & & & & \\ \hline

Multi-task & & & & & & \\
regression & NA &  NA & NA & NA & 0.5124 & 3.3659 \\ 
($E$ only) & & & & & & \\ \hline


     & 1 & 0.5362 & 0.5362 & 11.7542 & 0.5690 & 4.2050 \\ \cline{2-7}
Embedding $Y$      & 2 & 0.5362 & 0.4928 & 9.9780 & 0.4950 & 3.6555 \\ \cline{2-7}
+     & 5 & {\bf 0.6087} & {\bf 0.4058} & {\bf 9.2840} & {\bf 0.4516} & {\bf 3.3488} \\ \cline{2-7}
$k$NN & 10 & 0.5652 & 0.4783 & 10.1398 & 0.4622 & 3.4128 \\ \cline{2-7}
     & 15 & 0.5362 & 0.4928 & 10.4433 & 0.4798 & 3.4012 \\ \cline{2-7}
     & 20 & 0.4783 & 0.5652 & 10.9867 & 0.4813 & 3.4746 \\ \hline \hline

  
  & 1 & {\bf 0.6087} & 0.4783 & 10.9306 & 0.5515 & 4.3547 \\ \cline{2-7}
Pairwise $Y$  & 2 & 0.5362 & 0.5072 & 10.9274 & 0.5095 & 3.9330 \\ \cline{2-7}
+  & 5 & 0.5507 & {\bf 0.4638} & {\bf 10.4720} & 0.4935 & 3.6824 \\ \cline{2-7}
$k$NN  & 10 & 0.5072 & 0.5072 & 10.7297 & 0.4912 & {\bf 3.5969} \\ \cline{2-7}
  & 15 & 0.5217 & 0.4928 & 10.6659 & {\bf 0.4889} & 3.6277 \\ \cline{2-7}
  & 20 & 0.4638 & 0.5507 & 10.5957 & {\bf 0.4889} & 3.6576 \\ \hline \hline

	& 1 & {\bf 0.6087} & {\bf 0.4493} & 11.4919 & 0.5728 & 4.2644 \\ \cline{2-7}
Pairwise $E$	 & 2 & 0.4928 & 0.5072 & 9.7964 & 0.5072 & 3.7131 \\ \cline{2-7}
+	 & 5 & 0.5507 & {\bf 0.4493} & {\bf 9.6680} & 0.4767 & 3.4489 \\ \cline{2-7}
$k$NN	 & 10 & 0.5507 & {\bf 0.4493} & 9.9089 & 0.4897 & 3.4294 \\ \cline{2-7}
	 & 15 & 0.4928 & 0.5072 & 10.1360 & 0.4844 & {\bf 3.4077} \\ \cline{2-7}
	 & 20 & 0.4928 & 0.5072 & 10.0589 & {\bf 0.4760} & 3.3877 \\ \hline \hline

 
 & 1 & {\bf 0.6522} & {\bf 0.3913} & 10.4714 & 0.5431 & 4.0833 \\ \cline{2-7}
Pairwise $Y$\&$E$ & 2 & 0.5362 & 0.4783 & {\bf 10.0081} & 0.4882 & 3.6610 \\ \cline{2-7}
+ & 5 & 0.5652 & 0.4638 & 10.0519 & {\bf 0.4622} & {\bf 3.4735} \\ \cline{2-7}
$k$NN & 10 & 0.5072 & 0.5217 & 10.3872 & 0.4653 & 3.4786 \\ \cline{2-7}
 & 15 & 0.5072 & 0.5217 & 10.7218 & 0.4737 & 3.4955 \\ \cline{2-7}
 & 20 & 0.4493 & 0.5797 & 10.8590 & 0.4790 & 3.5027 \\ \hline 
\end{tabular}
\end{tiny}
\end{table*}

\subsection{Melanoma}
\ignore{
For this dataset, we use the same approaches we used for the AADB
dataset with a few modifications. 
}

We use all the approaches proposed in Section \ref{jointpairwise} to
obtain results for the Melanoma dataset: (a) simple regression
baselines for predicting $Y$ and for predicting $E$, 
(b) multi-task classification to
predict $Y$ and $E$ together, 
(c) $k$NN using embeddings from the
simple regression network ($Y$), and
from the baseline $E$ network, 
(d) $k$NN using embeddings optimized
for pairwise loss using $Y$, and using $E$.
We do not obtain
embeddings using weighted pairwise loss with $Y$ and $E$ because there
is a one-to-one map from $E$ to $Y$ in this dataset. 

The networks used a modified PyTorch
implementation of AlexNet for fine-tuning \citep{KrizhevskySH2012}. 
We simplified the fully connected layers for the regression variant of
AlexNet to 1024-ReLU-Dropout-64-$n$, where $n=1$ for predicting $Y$,
and $n=11$ for predicting $E$. 
We used cross-entropy losses.

In the multi-task case for predicting
$Y$ and $E$ together, the convolutional layers were shared and two
separate sets of fully connected layers with $1$ and $11$ outputs were
used. The multi-task network used a weighted sum of regression losses
for $Y$ and $E$: $\text{loss}_Y + \lambda \text{loss}_E$. All these
single-task and multi-task networks were trained for $100$ epochs with
a batch size of 64. The embedding layer that provides the
$64-$dimensional output had a learning rate of $0.01$, whereas all
other layers had a learning rate of $0.001$. For training the
embeddings using pairwise losses, we used $100,000$ pairs chosen from
the training data, and optimized the loss for $15$ epochs. 
The hyper-parameters
$(m_1,m_2)$ were set to $(0.75, 0.75)$, and were chosen to based on
the validation set performance. 
For the loss
(\ref{equation-loss-x-y}), $a$ and $b$ were said to be neighbors if
$y_a = y_b$ and non-neighbors otherwise.  For the loss
(\ref{equation-loss-x-e}), $a$ and $b$ were said to be neighbors if
$z_a = z_b$ and non-neighbors $y_a \neq y_b$. The pairs where $z_a
\neq z_b$, but $y_a = y_b$ were not considered. 

The left side of Table~\ref{table-Olf-Mel} provides accuracy numbers for $Y$ and $E$ using the proposed approaches. Numbers in bold are the best for a metric among an algorithm.  The $Y$ and $E$ accuracies for multi-task and $k$NN approaches are better than that the baselines, which clearly indicates the value in sharing information between $Y$ and $E$. The best accuracy on $Y$ is obtained using the Pairwise $E$ + $k$NN approach, which is not surprising since $E$ contains $Y$ and is more granular than $Y$. Pairwise $Y$ + $k$NN approach has a poor performance on $E$ since the information in $Y$ is too coarse for predicting $E$ well.

\subsection{Olfactory}
Since random forest was the winning entry on this dataset \citep{olfs}, we used a random forest regression to pre-select $200$ out of $4869$ features for subsequent modeling. From these $200$ features, we created a base regression network using fully connected hidden layer of 64 units (embedding layer), which was then connected to an output layer. No non-linearities were employed, but the data was first transformed using $\log10(100+x)$ and then the features were standardized to zero mean and unit variance. Batch size was 338, and the network with pairwise loss was run for $750$ epochs with a learning rate of $0.0001$.  For this dataset, we set $(c_1,c_2,c_3,c_4,m_1,m_2,w)$ to $(10,20,0.0272,0.0272,0.25,0.25,1.0)$. The parameters were chosen to maximize performance on the validation set.

The right side of Table~\ref{table-Olf-Mel} provides accuracy numbers in a similar format as the left side. The results show, once again, improved $Y$ accuracy over the baseline for Pairwise $Y$ + $k$NN and Pairwise $Y$ \& $E$ + $k$NN and corresponding improvement for MAE for $Y$. Again, this performance improvement can be explained by the fact that the predictive accuracy of $Y$ given $E$ using the both baselines were $0.8261$, with MAEs of $0.1739$ and $3.4154$ ($4.0175$ for RF) for Discretized and Continuous, respectively. Once again, the accuracy of $E$ varies among the 3 $k$NN techniques with no clear advantages. The multi-task linear regression does not perform as well as the Pairwise loss based approaches that use non-linear networks.


\section{Discussion}

\citet{ted-vision2019} discuss the additional labor required for
providing training explanations.  Researchers~\citep{zaidan-eisner:2008:gen,DBLP:journals/corr/ZhangMW16,McDonnel16why-relevant} have
quantified that for an expert SME, the
time to add labels and explanations is often the same as just adding
labels and cite other benefits, such
as improved quality and consistency of the resulting training data set.
Furthermore, in some instances, the $k$NN instantiation of TED may require no extra labor. For example, when embeddings are used as search criteria for 
evidence-based predictions of queries, end users will, on average, naturally interact with search results that are similar to the query in explanation space. This query-result interaction activity inherently provides similar and dissimilar pairs in the explanation space that can be used to refine an embedding initially optimized for the predictions alone. This reliance on relative distances in explanation space is also what distinguishes this method from multi-task learning objectives, since absolute labels in explanation space need not be defined.

\section{Conclusion} \label{sec-conc}
The societal demand for ``meaningful information'' on 
automated decisions has sparked significant research in AI explanability.
\ignore{
This paper suggests a new paradigm for providing explanations from machine learning algorithms.  This new approach is particularly well-suited for explaining a machine learning prediction when all of its input features are inherently incomprehensible to humans, even to deep subject matter experts. 
The approach augments training data collection beyond features and labels to also include elicited explanations.  Through this simple idea, we are not only able to provide useful explanations that would not have otherwise been possible, but we are able to tailor the explanations to the intended user population by eliciting training explanations from members of that group.

This paper has explored several instantiations of the \ted framework for teaching explanations to the machine learning process.
There are many possible instantiations for this proposed paradigm of teaching explanations.  We have}
This paper 
describes two novel instantiations of the \ted framework. The first learns feature embeddings using labels and explanation similarities in a joint and aligned way to permit neighbor-based explanation prediction. The second uses labels and explanations together in a multi-task setting. 
We have demonstrated these two instantiations on a publicly-available olfactory pleasantness dataset \citep{olfs} and Melanoma detection dataset \citep{codella2018skin}. We hope this work will inspire other researchers to further enrich this paradigm.

\bibliography{IEEEabrv,ted,ExAbsent}
\bibliographystyle{icml2019}

\end{document}